\pdfoutput=1
\documentclass[11pt]{article}

\usepackage[review]{acl}

\usepackage{times}
\usepackage{latexsym}

\usepackage[T1]{fontenc}
\usepackage[utf8]{inputenc}
\usepackage{microtype}
\usepackage{CJKutf8}
\usepackage{cite}
\usepackage{amsmath,amssymb,amsfonts}
\usepackage{algorithmic}
\usepackage{graphicx}
\usepackage{textcomp}
\usepackage{pdfpages}
\usepackage{booktabs}

\title{SLFNet: Generating Semantic Logic Forms from Natural Language Using Semantic Probability Graphs}

\author{ \bf
Hao Wu$^1$,
Fan Xu$^1$ \\ \\
$^1$School of Computer Science and Technology, \\
University of Science and Technology of China
}

\begin{document}
\nolinenumbers

% \maketitle
{\makeatletter\acl@finalcopytrue
  \maketitle
}

\begin{abstract}

Building natural language interfaces typically uses a semantic parser to parse the user's natural language and convert it into structured \textbf{S}emantic \textbf{L}ogic \textbf{F}orms (SLFs). The mainstream approach is to adopt a sequence-to-sequence framework, which requires that natural language commands and SLFs must be represented serially. Since a single natural language may have multiple SLFs or multiple natural language commands may have the same SLF, training a sequence-to-sequence model is sensitive to the choice among them, a phenomenon recorded as "order matters". To solve this problem, we propose a novel neural network, SLFNet, which firstly incorporates dependent syntactic information as prior knowledge and can capture the long-range interactions between contextual information and words. Secondly construct semantic probability graphs to obtain local dependencies between predictor variables. Finally we propose the Multi-Head SLF Attention mechanism to synthesize SLFs from natural language commands based on Sequence-to-Slots. Experiments show that SLFNet achieves state-of-the-art performance on the ChineseQCI-TS and Okapi datasets, and competitive performance on the ATIS dataset.

\end{abstract}

\section{Introduction}

\begin{figure}[t]
	\centering
	\begin{center}
	\includegraphics[width=0.9\linewidth]{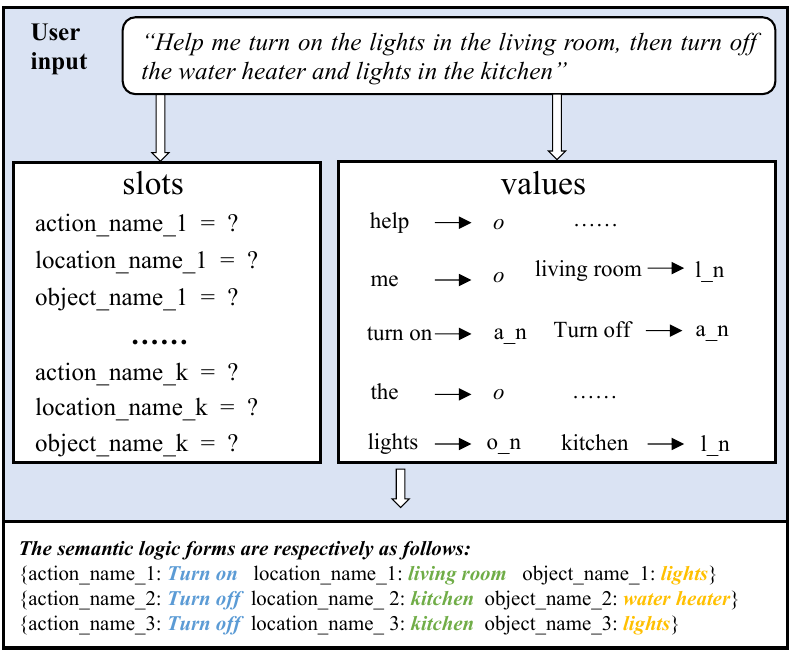}
	\caption{A natural language example from the dataset ChineseqCI-TS. In the framework of semantic parsing of natural language commands, semantic slots are defined as three types "$ALO$", where "$A$" represents the $action\_name$ slot; "$L$" stands for slot $location\_name$; "$O$" indicates the $object\_name$ slot; The $k$ at the end of the slot name represents the number of "$ALO$" groups that the natural language command may contain. For values predictions, this is a sequence labeling problem, where lower-case "$o$" means that the token does not belong to either of the "$ALO$" slots, where "turn on" fills in the $action\_name\_1$ slot, and the prediction details are described in the model description.}
	\label{fig:Real intent}
	\end{center}
\end{figure}

Semantic parsing (\citealp{aghajanyan-etal-2020-conversational, liang-etal-2013-learning, Berant2013SemanticPO, reddy-etal-2016-transforming, cheng-etal-2017-learning}) has been a hot research topic in the fields of human-computer interaction (\citealp{manaris1998natural, ogden1997using}), information retrieval (\citealp{smeaton1999using, feldman1999nlp}) and natural language understanding (\citealp{allen1988natural, dong2019unified}) for decades. The goal is to automatically translate unstructured natural language into structured semantic representations (\citealp{Jie2014MultilingualSP}) that machines can understand, such as logical forms (\citealp{10.5555/3020336.3020416}), abstract meaning representations (\citealp{pourdamghani-etal-2016-generating,koller-etal-2019-graph,xu2023few}), and structured queries (e.g., SQL (\citealp{Xu2017SQLNetGS,zhongSeq2SQL2017}), WebAPI (\citealp{10.1145/3132847.3133009,10.1145/3209978.3210013,Hosseini2021CompositionalGF}), etc.) to interact with data and services. However, due to the limitations of natural language understanding, scientific generalization, extensibility and interpretability, the early construction of natural language interfaces for data did not yield the desired results. In recent years, there has been a significant resurgence of natural language understanding systems in the form of virtual assistants (\citealp{10.1007/978-1-4614-8280-2_1, 7846294}), conversational systems (\citealp{klopfenstein2017rise}), semantic parsing, machine Q\&A (\citealp{Chen2021GeoQAAG}), and procedural generation systems (\citealp{summerville2018expanding}).

\section{Semantic Probability Graphs}

Different from previous semantic analysis models, we propose a prediction model based on semantic probabilistic graphs(\citealp{2104.06378}), as shown in Figure 1, which combines semantic parsing and probabilistic graph theory to graphically represent the dependency relationship between natural language commands and key tokens, i.e., when we make a prediction for one of the slots~\citealp{wu2024earthfarsser,xu2024revisiting,wu2024dynst,wangnuwadynamics,wu2024spatio,wu2023spatio,wang2024modeling,wu2023pastnet}, we only condition on those slots it depends on, thus maintaining the independence between different groups of SLFs.  The selection-sensitivity problem that arises in the sequence-to-sequence process is fundamentally avoided. The structure of our semantic probability graph is highly consistent with the syntax of the SLF form, so that SLFNet only needs to predict the number k of SLFs contained in the semantic probability graph and their corresponding slot values, without outputting both the syntax and the content of the SLFs.

All the slot values to be predicted in the semantic probability graph are represented by small boxes, and directed edges are used to describe the dependencies between the predicted values. For example, the box for $Location_2$ has two directed edges from NLC and $Action_2$ . These two directed edges indicate that the preconditions for the prediction of $Location_2$ depend on NLC and $Action_2$.This process of NL-to-SLF can be viewed as a causal inference problem on this semantic probabilistic graph. To synthesize SLFs using semantic probability graphs, we design SLFNet neural networks using three techniques, namely the Seq-to-Slots method, the Dependency-Fused BiLSTM coding model, and the Multi-Head SLF Attention mechanism. The details are described in Section 3.

\begin{figure*}[t]
	\centering
	\begin{center}
		\includegraphics[width=1.0\linewidth]{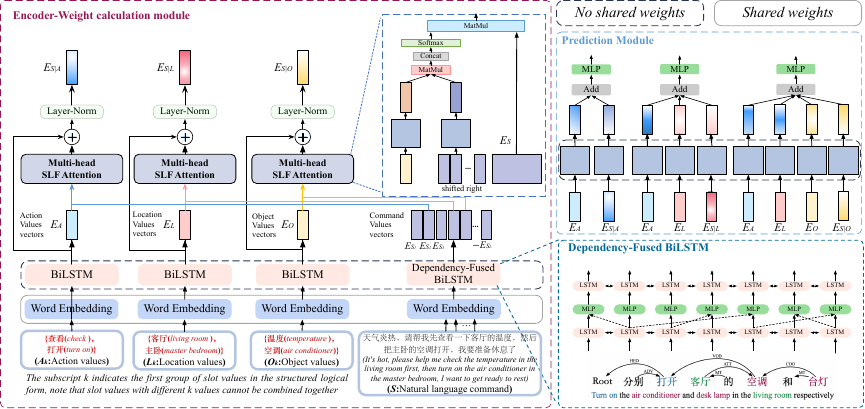}
	\caption{SLFNet neural network architecture diagram includes three main modules: (a). Encoder-Weight calculation module; (b). Denpendency-Fused BiLSTM module; (c). Prediction Module.}
	\label{fig:Real intent}
	\end{center}
\end{figure*}

\section{SLFNet}
The above semantic parsing problem is decomposed into three subtasks.(1) multi-classification task: identify how many groups of SLFs are contained in natural language commands, i.e. $k \in (0,i+1)$; (2) slot-filling task: predict the values of slots in the semantic probability graph, i.e. the values corresponding to Action, Location, and Object; (3) logical inference(\citealp{chen-etal-2017-enhanced}): form optimal permutations based on the first two tasks to synthesize SLFs. The SLFNet network structure diagram is shown in Figure 3.

\subsection{Seq-to-Slots based prediction method using Multi-Head SLF Attention} 

{\bfseries Sequence-to-Slots.} All the tokens in SLFs constitute a subset of the tokens in the NLC. Therefore, we don't need to generate all the tokens in SLFs at once. We can directly predict which tokens in the NLC should be in the SLFs. This article is called the Sequence-to-slots prediction. The following is an example of predicting \textbf{Action} slot values.

In particular, we calculate the probability $P_{S \text {-action }}(A|S)$ ,where $A$ is the slot value in $Action$ attribute, $S$ is the input NLC, and the formula is shown below.
\begin{equation}
P_{S \text {-action }}(A \mid S)=\operatorname{Sigmoid}\left(W_{a}^{T} E_{A}+W_{s}^{T} E_{S}\right)
\end{equation}
where $E_{A} \in \mathbb{R}^{d}$ and $E_{S} \in \mathbb{R}^{d}$ are the Action slot values and the embedding of NLC, respectively.$W_{a} \in \mathbb{R}^{d}$ and $W_{s} \in \mathbb{R}^{d}$ are two trainable weights. The embedding $E_{A}$ is computed as the hidden state of the Action slot value on the Bi-LSTM, and the embedding $E_{S}$ is computed as the hidden state of the NLC sequence on the Dependency-Fused BiLSTM. As shown in Figure 3, the Bi-LSTMs used to encode slot values and natural language command do not share weights between them.

It is worth noting that we want to find the token most relevant to the predicted label in the NLC. to capture this information, we first need to calculate the weights of each token in the NLC, and then weight these weights to sum up.  Also intuition suggests that the weight of each token is also related to the syntactic structure of the whole natural language command, so the use of Dependency-Fused BiLSTM in the encoding part of the NLC is one of the reasons. In summary we propose a new attention mechanism Multi-Head SLF Attention.

{\bfseries Multi-Head SLF Attention.} We design a new Multi-Head SLF Attention mechanism to compute $E_{S \mid A}$. First, we assume that $E_{S}$ is a matrix of size $d \times L$, where $L$ represents the length of the NLC and the i-th column of the NLC represents the output of the Dependency-Fused BiLSTM hidden layer corresponding to the i-th token in the NLC, denoted by  $E_{S_i}$, whose dimension is d.

We calculate the attention weight value $a_w$ for each token in NLC. Multi-Head SLF Attention mechanism block calculation steps are shown in Figure 3. First we consider the input $E_A$ as \textbf{query}, $E_S$ as key and Value respectively, then calculate the similarity of $E_A$ and $E_{S_i}$ as weights, and finally perform concat operation on all the weights, normalize them and multiply them with Value. The specific calculation is as follows:
\begin{equation}
\begin{gathered}
v_{i}=\left(W^{Q} E_{A}\right)^{T} W^{K} E_{S_{i}} \\
v=\operatorname{Concat}\left(v_{1}, \ldots, v_{i}\right) \\
a_{w}=\operatorname{soft} \max (v) \\
E_{S \mid A}=E_{S} a_{w}
\end{gathered}
\end{equation}
In particular, inspired by (\citealp{NIPS2017_3f5ee243,xu2023exploring}), we believe that the query, key, and value are projected into $h$ different SLF Attentions to calculate $h$ output weights, from $E_{S \mid A}^{1}$ to $E_{S \mid A}^{h}$ , and then concat, and finally The final output $E_{S \mid A}^{H}$ is obtained by passing it to the linear layer, where the dimension of $W^h$ is $(L \times 1)$ . The calculation formula is shown below:
\begin{equation}
E_{S \mid A}^{H}=\operatorname{Concat}\left(E_{S \mid A}^{1}, \ldots, E_{S \mid A}^{h}\right) W^{h}
\end{equation}
We can replace $E_S$ in Equation (1) with  $E_{S \mid A}$ and build a new Seq-to-Slots prediction model based on Multi-Head SLF Attention.
\begin{equation}
P_{S \text {-action }}(A \mid S)=\operatorname{Sigmoid}\left(W_{a}^{T} E_{A}+W_{s}^{T} E_{S \mid A}\right)
\end{equation}

But the specific prediction model building we will detail in subsection 3.3, which has a slightly different form than here, but the basic idea is the same, and we also demonstrate in the experimental section that our Multi-Head SLF Attention effectively improves the prediction accuracy of the model.

\subsection{Slot values prediction} 

{\bfseries SLF groups with multiple classifications.} The number of SLFs contained in the NLC needs to be identified. The number of SLFs contained in NLC is limited, and the upper limit is set to i. The essence is a multi-category problem, and a multi-category classification model(\citealp{zhou-etal-2016-attention}) needs to be built, with the following equation.

\begin{equation}
\begin{aligned}
&P_{\text {class }}(k \mid S)=\operatorname{soft} \max \left(W_{1}^{\text {class }}\right. \\
&\tanh \left(W_{2}^{S \mid A} E_{S \mid A}\right.+W_{3}^{S \mid L} E_{S \mid L} \\
&\left.\left.+W_{4}^{S \mid O} E_{S \mid O}\right)\right)_{i}
\end{aligned}
\end{equation}

\begin{equation}
\boldsymbol{k}=\operatorname{argmax}\left(P_{\text {class }}(k \mid S)\right)
\end{equation}

Where $W_{1}^{\text {class }}$, $W_{2}^{S \mid A}$, $W_{2}^{S \mid L}$, $W_{2}^{S \mid O}$ are trainable parameter matrices with dimension sizes of $(i+1) \times d$, $(d \times d)$, $(d \times d)$, $(d \times d)$, $(d \times d)$ respectively. $\operatorname{softmax}(\ldots \ldots)_{i}$ means outputting the $i-th$ dimension, i.e., indicating the $i-th$ category. Finally SLFNet chooses the group number $k$ that maximizes $P_{\text {class }}(k \mid S)$, thus obtaining the number of SLFs contained in the NLC. 

{\bfseries Slot values filled.} 
In this subsection, we describe in detail the process of SLFNet to predict \textbf{Action}, \textbf{Location} and \textbf{Object} slot values. The specific prediction method is shown below.

\begin{itemize}
    \item \textbf{ACTION Slot values predicted.}
    It is worth noting that the example we used when we introduced the Seq-to-Slots method and the Multi-Head SLF Attention mechanism in subsection 3.2 was predicting \textbf{Action} values. But it is not perfect enough, and predicting the slot value of \textbf{Action} is different from the slot value method of \textbf{Location} and \textbf{Object}. The specific difference is that we calculate all the \textbf{Action} slots contained in the natural language command at once, if k=3 in the command, we predict $A_1$, $A_2$ and $A_3$ at once, and then fill them into the corresponding $Action$ slots in the semantic probability graph. However, when predicting $Location$ and $Object$, we adopt the approach of predicting $L_1$, $O_1$, then $L_2$, $O_2$, and finally $L_3$, $O_3$, because we found that this is more conducive to SLFNet to complete logical inference and produce better logical combinations during our experiments.
    We give the equation to calculate the probability $P_{S \text {-action }}(A \mid S)$ in subsection 3.3, but in fact, we follow the prediction module in Figure 3 to obtain the final model for predicting the Action slot values.
\begin{equation}
\begin{aligned}
&P_{S-a c t i o n}(A \mid S)=\operatorname{Sigmoid}(z) \\
&z=\left(W_{A}^{\mathrm{ac}}\right)^{T} \tanh \left(W_{a}^{a c} E_{A}+W_{s}^{\mathrm{ac}} E_{S \mid A}\right)
\end{aligned}
\end{equation}
where $W_{A}^{\mathrm{ac}}$, $W_{a}^{\mathrm{ac}}$, and $W_{s}^{\mathrm{ac}}$ are trainable parameter matrices with $d \times 1$, $d \times d$, and $d \times d$ dimensional sizes, respectively.
    \item \textbf{Location Slot values predicted.}
    For the \textbf{LOCATION} slot, we need to predict all tokens from the NLC. these tokens are a complete subset of all tokens in the NLC. The encoding part is the same as the method used to predict the \textbf{ACTION} slots, i.e., the Encoder-Weight calculation module part of the figure 3, and the decoding part uses Pointer Networks(\citealp{NIPS2015_29921001}) based on the Muilt-Head SLF Attention mechanism to compute which tokens belong to the \textbf{LOCATION} slots.The output of each token encoded in the NLC is $E_{s_i}$. The probability of predicting the \textbf{LOCATION} slot values from the prediction module of Fig. 3 can be obtained as:
\begin{equation}
\begin{aligned}
&P_{S \text {-location }}(i \mid S, \text { action, location }) \\
&=\operatorname{soft} \max \left(\left(W_{L}^{\text {lo }}\right)^{T}\right. \\
&\left.\tanh \left(W_{1}^{\text {lo }} E_{s_{i}}+W_{2}^{\text {lo }} E_{A}+W_{3}^{\text {lo }} E_{S \mid \mathrm{L}}\right)\right)_{i}
\end{aligned}
\end{equation}
where $W_{L}^{\mathrm{lo}} \in \mathbb{R}^{d \times 1}$,$W_{1}^{\mathrm{lo}} \in \mathbb{R}^{d \times d}$,$W_{2}^{\mathrm{lo}} \in \mathbb{R}^{d \times d}$,$W_{3}^{\mathrm{lo}} \in \mathbb{R}^{d \times d}$ are trainable matrices, where $i \in\{1, \ldots, L\}$.

 \item \textbf{Object Slot values predicted.}For the prediction of $OBJECT$ slot values, we know from the semantic probability graph that the dependent prerequisites are the predicted values of the $ACTION$ and $LOCATION$ slots and the NLC. Consistent with the prediction method for $LOCATION$ slot values, the calculation formula is:
 \begin{equation}
\begin{aligned}
&P_{S-o b j e c t}(i \mid S, \text { action,location, Object })= \\
&\operatorname{soft} \max \left(\left(W_{O}^{o b}\right)^{T}\right. \tanh \left(W_{1}^{o b} E_{s_{i}}+W_{2}^{o b} E_{A}\right. \\
&\left.\left.+W_{3}^{o b} E_{L}+W_{4}^{o b} E_{S \mid \mathrm{O}}\right)\right)_{i}
\end{aligned}
\end{equation}
where $W_{O}^{\mathrm{ob}} \in \mathbb{R}^{d \times 1}$,$W_{1}^{\mathrm{ob}} \in \mathbb{R}^{d \times d}$,$W_{2}^{\mathrm{ob}} \in \mathbb{R}^{d \times d}$,$W_{3}^{\mathrm{ob}} \in \mathbb{R}^{d \times d}$,
$W_{4}^{\mathrm{ob}} \in \mathbb{R}^{d \times d}$are trainable matrices, where $i \in\{1, \ldots, L\}$.
\end{itemize}

\section{Experiment}
\label{sec:ourExper}

\subsection{Data} 

We evaluated our model on the $IEEE2021$ dataset $ChineseQCI-TS$ (\citealp{9440396}). The dataset was expanded to 21900 entries by crowdsourcing using the method in SCADA-NLI (\citealp{9440396}). On this dataset, we evaluated in the ratio of training set, validation set, and test set 6:2:2. In addition, we evaluated our model on the datasets WikiSQL(\citealp{zhongSeq2SQL2017}) and Okapi(\citealp{2112.05209}) to verify the scalability of the model. 

\subsection{Baselines}

\begin{table*}[!t]  \caption{\textbf{The overall results of ChineseQCI-TS, where SLFNet represents the original network, SLFNet+SA represents the neural network incorporating SLF Attention mechanism, SLFNet+EM represents the neural network incorporating dependent syntactic information coding, and SLFNET+EM+SA represents the neural network incorporating dependent syntactic information coding and SLF Attention mechanism.}}
 \label{table1}
 \centering
 \begin{tabular}{lllllllll}  
\toprule   
Model&	$Acc_{dev}$&	$Pre_{dev}$&	$Rec_{dev}$&	$F_{dev}$&	$Acc_{test}$&	$Pre_{test}$&	$Rec_{test}$&	$F_{test}$ \\
\midrule   
Vanilla Seq2seq&	53.2\%&	52.1\%&	55.7\%&	53.8\%&	54.3\%&	55.7\%&	56.3\%&	56.0\% \\
Modular Seq2Seq	&66.4\%&	63.2\%&	65.5\%&	64.9\%&	66.4\%&	64.5\%&	65.3\%	&64.9\% \\
BERT+Copy&	67.5\%&	71.2\%&	72.2\%&	71.7\%&	68.5\%&	73.2\%&	73.3\%&	73.3\% \\
SLFNet&	59.3\%&	60.1\%&	61.2\%&	60.6\%&	60.1\%&	64.4\%&	63.9\%&	64.0\% \\
SLFNet+SA&	63.2\%&	64.1\%&	63.9\%&	64.0\%&	65.4\%&	66.2\%&	67.3\%&	66.7\% \\
SLFNet+EM&	64.5\%&	66.9\%&	67.1\%&	70.0\%&	65.7\%&	67.6\%&	67.4\%&	67.5\% \\
SLFNet+EM+SA&	78.4\%&	80.1\%&	82.6\%&	81.3\%&	79.7\%&	82.4\%&	84.9\%&	83.4\% \\
\bottomrule  
\end{tabular}
\end{table*}

In the experiment, SLFNet will be compared with the following baselines.

{\bfseries Vanilla Seq2seq}: the general seq2seq model(\citealp{10.5555/2969033.2969173}) with GRU(\citealp{cho-etal-2014-learning}) method.

{\bfseries Modular Seq2Seq}: A novel modular sequence-to-sequence model Source for the paper (\citealp{10.1145/3209978.3210013}).

{\bfseries BERT+Copy}:This is a generic and frequently used model in the field of semantic parsing, which consists of a pre-trained BERT encoder (\citealp{devlin-etal-2019-bert}) and an autoregressive transformer decoder (\citealp{NIPS2017_3f5ee243}) that focuses on the output of the encoder. For repetitive entity information, an attention-based COPY mechanism(\citealp{gu-etal-2016-incorporating}) is adopted for replication extraction.

\subsection{Results and Discussion}

The results of the comparison experiments are shown in Table 1. Compared with the common Seq2Seq model and its variants, the SLFNet+EM+SA model is superior in overall performance. Its accuracy achieves about 80\% accuracy on both the validation and training sets. It is significantly ahead of the base model. At the same time, for SLFNet itself, we found that the Dependency-Fused BiLSTM coding model and Multi-Head SLF Attention mechanism also outperformed the original SLFNet model by a wide margin. This demonstrates the effectiveness of both approaches.

\section{Conclusion}
\label{sec:Conclusion}
In this paper, we propose an approach to handle the NL2SLF task, namely SLFNet. we observe that all models employing sequence-to-sequence suffer from sequence selection sensitivity when order does not matter. We propose a novel neural network architecture, SLFNet, which firstly fuses the neural network model of dependent syntactic trees capable of capturing contextual information. Secondly, semantic probability maps are constructed to capture local dependencies between predictor variables so that only the previous predictions on which it depends need to be considered when predicting a variable. Overall, we observe a significant improvement of our SLFNet architecture over the traditional Seq2seq approach and its variants, ranging from 11\% to 25\%. This suggests that our approach can effectively address the above problem, shedding new light on novel solutions to the structure generation problem.

\bibliographystyle{acl_natbib}
\bibliography{custom}

\appendix
\section{Example Appendix}

\label{sec:appendix}

\subsection{The derivation process of  Denpendency-Fused BiLSTM}
The architecture with a 2-layer BiLSTM shown in Figure 3 can efficiently encode the grandchild dependencies because the input representation encodes the parent information and the interaction function further propagates the grandfather information. This propagation allows for long-range interactions that the model captures indirectly from the dependencies between words in the sentence. The hidden layer state $\mathbf{H}^{(l+1)}$ of the $(l+1)$-th layer can be computed from the hidden state $\mathbf{H}^{l}$ of the previous layer as follows:

\begin{equation}
\begin{aligned}
\mathbf{H}^{(l+1)} &=\operatorname{BiLSTM}\left(f\left(\mathbf{H}^{(l)}\right)\right) \\
\mathbf{H}^{(l)} &=\left[\mathbf{h}_{1}^{(l)}, \mathbf{h}_{2}^{(l)}, \cdots, \mathbf{h}_{n}^{(l)}\right] \\
f\left(\mathbf{H}^{(l)}\right) &=\left[g\left(\mathbf{h}_{1}^{(l)}, \mathbf{h}_{p_{1}}^{(l)}\right), \cdots, g\left(\mathbf{h}_{n}^{(l)}, \mathbf{h}_{p_{n}}^{(l)}\right)\right]
\end{aligned}
\end{equation}
$g\left(\mathbf{h}_{i}^{(l)}, \mathbf{h}_{p_{i}}^{(l)}\right)$ denotes the interaction function between the hidden states at positions $i$ and $p_i$ under the dependent edge $(x_{p_i},x_i)$. The number of layers $L$ can be chosen according to the performance of the development set.
\subsection{Metrics}

To evaluate our approach, we use the metrics as below:

{\bfseries Accuracy:} The accuracy rate indicates the ratio of the number of correctly executed commandns to the number of all commandns executed, and in this paper, we use $C$ to denote the number of correctly executed commandns and $T$ to denote the total number of commandns. The formula is shown as follows.
\begin{equation}
\text { Accuracy }=\frac{C}{T}
\end{equation}

{\bfseries Precision:} The precision rate indicates the value of the correctly expressed slots and the proportion of the corresponding slots in the obtained SLFs results. The number of correctly predicted slot values is denoted by $L$, and all SLFs slot values are denoted by $P$. The calculation formula is as follows.
\begin{equation}
\text { Accuracy }=\frac{L}{P}
\end{equation}

{\bfseries Recall:}Recall is the number of slot values correctly expressed in SLFs obtained by the prediction model for all slot values R contained in the original command. The formula is as follows.
\begin{equation}
\text { Accuracy }=\frac{L}{R}
\end{equation}
{\bfseries F-Scores:}F-score is the summed average of precision rate and recall rate.
\begin{equation}
\text { F-score }=\frac{2 \times \text { Precision } \times \text { Recall }}{\text { Precision }+\text { Recall }}
\end{equation}

\subsection{Training details}
Here, we give the details of the model training

\textbf{Input encoding model details.} Natural language commands and slot values are treated as a series of tokens. We use DDParser(\citealp{zhang2020practical}) to parse the sentences. Each token is represented as a single heat vector, and a word embedding vector is input before feeding them into a bidirectional LSTM. For this purpose, we use the GloVe(\citealp{pennington2014glove}) word embedding.

\textbf{Additional experiments.}
Here we add two experiments, the first one is to test the prediction accuracy of slot values. The second experiment is a migration task that will be performed.

\begin{figure}[t]
	\centering
	\begin{center}
		\includegraphics[width=1.0\linewidth]{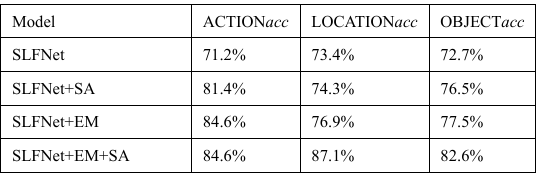}
	\caption{Accuracy of different slot values under SLFNet neural network.}
	\label{fig:Real intent}
	\end{center}
\end{figure}

\end{document}